\documentclass[10pt]{article}
\usepackage{amsmath,mathpazo,fullpage,amsthm,amssymb}
\usepackage[utf8]{inputenc}
\usepackage{amsmath,amssymb,amsfonts}
\usepackage{algorithmic}
\usepackage{graphicx}
\usepackage{textcomp}
\usepackage{xcolor}
\usepackage{booktabs}
\usepackage{tablefootnote}
\usepackage{subcaption}
\usepackage{dblfloatfix}
\usepackage{adjustbox}
\usepackage{colortbl}
\usepackage{hyperref} 
\usepackage{url}
\usepackage[numbers]{natbib}
\usepackage[load=prefixed]{siunitx}
\definecolor{Gray}{gray}{0.9}

\newcommand{\MeAPE}{\textsc{MeAPE}}
\newcommand{\AggPE}{\textsc{AggPE}}
\newcommand{\MAPE}{\textsc{MAPE}}
\newcommand{\PRMSE}{\textsc{\%RMSE}}
\newcommand{\MAE}{\textsc{MeAE}}

\title{Census-Independent Population Estimation using \\
Representation Learning
}

\begin{document}

\date{}

\author{Isaac Neal$^1$, Sohan Seth$^1\ast$, Gary Watmough$^1$, Mamadou S. Diallo$^2$\\
University of Edinburgh$^1$ and United Nations Children's Fund (UNICEF)$^2$\\
\texttt{\{ineal,sseth,gary.watmough\}@ed.ac.uk, mamsdiallo@unicef.org}
}

\maketitle

\begin{abstract}
Knowledge of population distribution is critical for building infrastructure, distributing resources, and monitoring the progress of  sustainable development goals. Although censuses can provide this information, they are typically conducted every ten years with some countries having forgone the process for several decades. Population can change in the intercensal period due to rapid migration, development, urbanisation, natural disasters, and conflicts. Census-independent population estimation approaches using alternative data sources, such as satellite imagery, have shown promise in providing frequent and reliable population estimates locally. Existing approaches, however, require significant human supervision, for example annotating buildings and accessing various public datasets, and therefore, are not easily reproducible. We explore recent representation learning approaches, and assess the transferability of representations to population estimation in Mozambique. Using representation learning reduces required human supervision, since features are extracted automatically, making the process of population estimation more sustainable and likely to be transferable to other regions or countries. We compare the resulting population estimates to existing population products from GRID3, Facebook (HRSL) and WorldPop. We observe that our approach 
matches the most accurate of these maps, and is interpretable in the sense that it recognises built-up areas to be an informative indicator of population.
\end{abstract}

\allowdisplaybreaks

\begin{figure*}[t]
  \includegraphics[width=\textwidth]{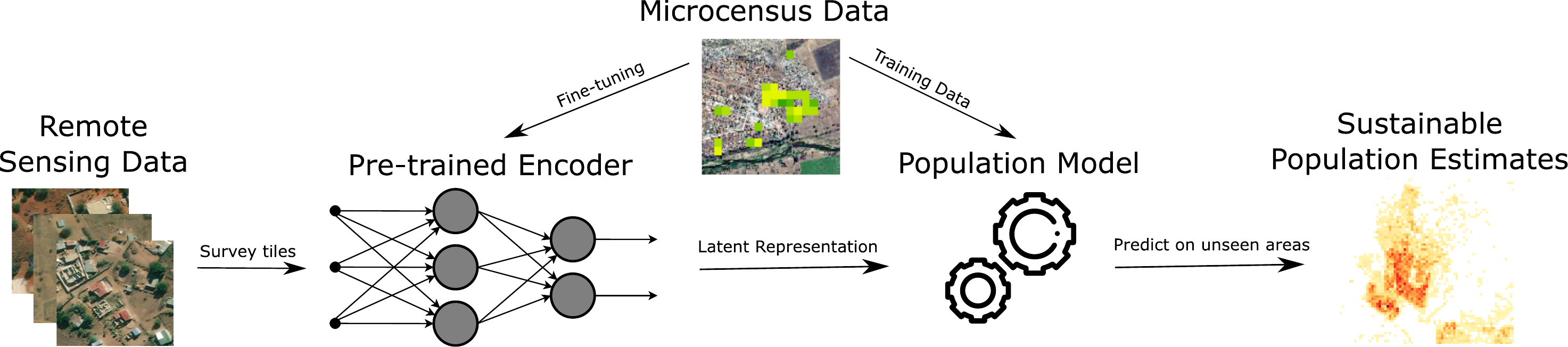}
  \caption{The figure illustrates our approach of sustainable census-independent population estimation, or SCIPE, using representation learning. Satellite images of surveyed grid tiles are mapped to vector representations through a pre-trained deep neural network, and a regression model is trained on the representation space to estimate population using microcensus. The pre-trained network can also be fine-tuned with microcensus to learn better representation indicative of population.}
  \label{fig:teaser}
\end{figure*}

\section{Introduction}
Accurate population maps are critical for planning infrastructure and distributing resources, including public health interventions and disaster relief, as well as monitoring well-being e.g. through the fulfillment of sustainable development goals \cite{robinson2017deep}. Censuses provide a `gold standard' population map by surveying every individual household nationally. However, this is an extremely expensive endeavor, and therefore they are typically conducted every ten years \cite{Shearmur10editorial}. During the intercensal period projections of the population are available at a coarser \emph{enumeration} level that is estimated either using birth, death, and migration rates \cite{robinson2017deep}, or through combining sample household surveys using small area estimation \cite{Shearmur10editorial}. Estimating finer-resolution intercensal population, e.g., over a \SI{100}{\m} grid, has received significant interest in the last decade, and several population maps, e.g. WorldPop \cite{linard2012population}, High-Resolution Settlement Layer (HRSL) \cite{hrsl,tiecke_mapping_2017} and GRID3 \cite{soton445265}, have been made available publicly to aid humanitarian initiatives. These approaches primarily use satellite imagery as a predictor, and can be broadly categorized as \textit{census-dependent} and \textit{census-independent} based on their use of census as response variable \cite{wardrop2018spatially}.

Census-independent population estimation approaches (\S\ref{sec:censusindpop}) using microcensus (survey data) are gaining prominence since they can improve the spatial and temporal resolution of census-dependent approaches (\S\ref{sec:deeppop}). However, finding census-independent methods that are sustainable and transferable, and informative data sources that are reliable and easy to procure remain active areas of research. Existing approaches rely heavily on hand-crafted features that often require manual curation, and the features used in modelling vary significantly between publications and countries where population is being estimated, making them less sustainable for repeated use, and less transferable to other regions and countries. For example, these approaches often use objects in satellite imagery, e.g., buildings and cars, \cite{engstrom_estimating_2020} and distribution of services, e.g., school density \cite{leasure_national_2020} and road networks \cite{wardrop2018spatially}, as indicators of population. Detecting building footprints usually requires manual annotation and curation while information on road networks and school density can be incomplete, unreliable, and difficult to procure.

Recent advances in \emph{representation learning}
have demonstrated that features, or \emph{representations}, automatically extracted from images through end-to-end training of a deep neural network can significantly improve the performance in many computer vision tasks in a sustainable manner by removing the need for hand-crafted features \cite{razavian_cnn_2014}. Additionally, \emph{transfer learning} can leverage features learned from a vast amount of annotated data from a separate task to improve performance on tasks lacking sufficient annotated data \cite{bengio_representation_2013}. Furthermore, \emph{explainable AI} has provided meaningful insight from these, so called `black box', models to explain the decisions made by them, enhancing their transparency \cite{linardatos_explainable_2020}. Representation learning can vastly simplify the problem of estimating population from satellite imagery by removing the need for handcrafted features, manual data procurement, and human supervision, thus improving the sustainability and transferability of the process. Additionally, transfer learning removes the need for large scale training data, allowing fine-tuning on limited microcensus with minimal computational resources. Finally, these methods provide interpretation of model outcome, promoting trust around the estimated population among the end-users.

We assess the utility of representation learning in census-independent population estimation from very-high-resolution
($\leq$\SI{5}{\m}) satellite imagery using a retrospective microcensus in two districts of Mozambique. To the best of our knowledge, we are the first to explore the potential of such approach, and in using both very-high (\SI{50}{\cm} spatial) resolution satellite imagery and microcensus in this manner. We observe that the proposed approach is able to produce a reliable medium-resolution (\SI{100}{\m}) population density map with minimal human supervision and computational resources. 
We find that this approach performs similar to the more elaborate approach of using building footprints to estimate population, and outperforms techniques using only public datasets to estimate population in our ROIs (Table~\ref{tab:results}). It also completely avoids manual annotation of satellite images and manual access to public datasets making it potentially more transferable to other countries using \emph{only very-high-resolution satellite imagery and gridded microcensus}. Additionally,  we observe that this approach learns to predict population in a reasonable manner by using built-up area as an indicator of population. We refer to this approach as \emph{Sustainable Census-Independent Population Estimation} or SCIPE (see Figure~\ref{fig:teaser} for an illustration), with our core motivation being developing population estimation methods that are easy to use, computationally efficient, and that can produce frequent and explainable population maps with associated uncertainty values. This will help humanitarian organizations extrapolate local microcensus information to regional level with ease, and provide more confidence in using the estimated population map in conjunction with existing ones. 

\begin{table}[t]
\centering
\resizebox{\linewidth}{!}{
    \begin{tabular}{@{}lllllll@{}}
        \toprule
        & & & & \multicolumn{3}{c}{Performance} \\
        \cmidrule(l){5-7}
        ROI & Input Resolution & Output Resolution & Input Data Cost & Validation & \MeAPE & $R^2$ \\
        \midrule
        \rowcolor{Gray}
        \begin{tabular}[c]{@{}l@{}}Kano and Kaduna,\\ Nigeria \cite{weber2018census}\end{tabular} & \SI{0.5}{\m} (Maxar) & \SI{90}{\m} & High (Maxar) & Survey (same region) & - & 0.98\\   
        Sri Lanka \cite{engstrom_estimating_2020} & \begin{tabular}[c]{@{}l@{}}\SI{10}{\m} (Object shape data)\\ 12-\SI{30}{\m} (Settlement Layer)\\
        \SI{750}{\m} (Night time Lights) \end{tabular} & Village level & \begin{tabular}[c]{@{}l@{}}Free (Landsat)\\ High (Maxar)\end{tabular} & Train / test split & 28 & 0.58 \\
        \rowcolor{Gray}
        Bo, Sierra Leone \cite{hillson_estimating_2019} & \SI{30}{\m} (Landsat) & City district level & Free (public data) & LOOCV & 11 & - \\
        Nigeria \cite{leasure_national_2020} & 
        \begin{tabular}[c]{@{}l@{}}\SI{0.5}{\m} (Maxar),\\ \SI{100}{\m} (WorldPop),\\ various (OSM school \\ density, household size)\end{tabular} & \SI{100}{\m} & \begin{tabular}[c]{@{}l@{}}Free (OSM,\\WorldPop)\\ High (Maxar)\end{tabular} & Train / test split & - & 0.26 \\
        \bottomrule
    \end{tabular}
}
\caption{Summary of recent literature on census-independent population estimation. NOTE: ROI is region of interest, LOOCV is leave one out cross validation, \MeAPE is median absolute percent error.}
\label{tab:bottomup}
\end{table}

\section{Background}
\label{sec:background}

In this section, we provide a detailed overview of the existing literature on census-independent population estimation (\S \ref{sec:censusindpop} and Table~\ref{tab:bottomup}) and the application of deep neural networks in intercensal census-dependent population estimation (\S \ref{sec:deeppop} and Table~\ref{tab:deeplearningpop}).

\subsection{Census-independent population estimation}
\label{sec:censusindpop}
\emph{Census-dependent population estimation}, also known as \emph{population disaggregation} or \emph{top-down estimation}, either uses census data to train a predictive model that can estimate population of a grid tile directly \cite{robinson2017deep}, or to train a model to estimate a weighted surface that can be used to distribute coarse resolution projected data across a finer resolution grid \cite{doupe2016equitable}. \textit{Census-independent population estimation}, also known as \textit{bottom-up estimation}, instead relies on locally conducted microcensuses to learn a predictive model that can estimate population at non-surveyed grid tiles. 

\textbf{Weber et al.}\cite{weber2018census} used very-high-resolution satellite imagery to estimate the number of under-5s in two states in northern Nigeria in three stages: first, by building a binary settlement layer at \SI{8}{\m} resolution using support vector machine (SVM) with ``various low-level contextual image features" \cite[\S 2.3]{weber2018census}
, second, by classifying ``blocks" constructed from OpenStreetMap data using ``a combination of supervised image segmentation and manual correction of errors" in 8 residential types (6 urban, 1 rural and 1 non-residential) \cite[\S 2.4]{weber2018census}
, and finally, by modelling population count of each residential type with separate log-normal distributions using microcensus. The predictions were validated against a separate survey from the same region, and were found to be highly correlated with this data \cite[\S 3.3]{weber2018census}.

\textbf{Engstrom et al.} \cite{engstrom_estimating_2020} used LASSO regularized Poisson regression and Random Forest models to predict village level population in Sri Lanka. The authors used a variety of remote sensing indicators at various resolutions as predictors, both coarser-resolution publicly available ones such as night time lights, elevation, slope, and tree cover, and finer-resolution proprietary ones such as built up area metrics, car and shadow shapefiles, and land type classifications \cite[Table 1]{engstrom_estimating_2020}. The authors observed that publicly available data can explain a large amount of variation in population density for these regions, particularly in rural areas, and the addition of proprietary object footprints further improved performance. Their population estimates were highly correlated with census counts at the village level \cite[Table 4]{engstrom_estimating_2020}.

\textbf{Hillson et al.} \cite{hillson_estimating_2019} explored the use of \SI{30}{m} resolution Landsat 5 thematic mapper (TM) imagery to estimate population densities and counts for 20 neighborhoods in the city of Bo, Sierra Leone. The authors used 379 candidate Landsat features generated manually,
which was reduced to 159 covariates through ``trial-and-error" \cite[p.~10]{hillson_estimating_2019}
and removal of highly correlated (Pearson's $\rho$ $>0.99$) pairs, and finally, an optimal regression model was learned using only 6 of these covariates \cite[Table~7]{hillson_estimating_2019}. 
These estimates were then validated through leave-one-out cross-validation on the districts surveyed. The approach estimated population density at the coarse neighborhood level with low relative error for most neighborhoods \cite[Table 10]{hillson_estimating_2019}.

\begin{table}[t]
\footnotesize
\centering
\resizebox{\linewidth}{!}{%
\begin{tabular}{@{}lllllllll@{}} 
\toprule
& & & & & \multicolumn{4}{c}{Performance} \\ \cmidrule(l){6-9} 
ROI & Input Shape & \begin{tabular}[c]{@{}l@{}}Input\\Resolution\end{tabular} & \begin{tabular}[c]{@{}l@{}}Output\\Resolution\end{tabular} & \begin{tabular}[c]{@{}l@{}}Input\\Data Cost\end{tabular} & Validation & \PRMSE & \MAPE & $R^2$ \\ \midrule 
\rowcolor{Gray}
Tanzania \cite{doupe2016equitable} & $32\times32\times8$ & \SI{250}{\m} (Landsat) & \SI{8}{\km} & Free & Train / test split & 51.5 & - & - \\
USA \cite{robinson2017deep} & $74\times74\times7$ & \SI{15}{\m} (Landsat) & \SI{1}{\km} & Free & \begin{tabular}[c]{@{}l@{}}Future census\end{tabular} & - & 49.8 & 0.94\\
\rowcolor{Gray}
India \cite{hu_mapping_2019} & $224\times224\times3$ & \begin{tabular}[c]{@{}l@{}}\SI{20}{\m} (Landsat)\\ \SI{20}{\m} (Sentinel-1 radar)\end{tabular} & \begin{tabular}[c]{@{}l@{}}\SI{4.5}{\km} \\ (Village level) \end{tabular} & Free & Train / test split & 24.3 & 21.5 & 0.93 \\ \addlinespace \bottomrule
\end{tabular}%
}
\caption{Summary of recent literature on deep learning driven census-dependent intercensal population estimation. NOTE: Input Resolution indicates the spatial resolution of imagery input to model (i.e. after reprojection and resizing). {\PRMSE} is percent root mean squared error, {\MAPE} is mean absolute percent error.}
\label{tab:deeplearningpop}
\end{table}

\textbf{Leasure et al.} \cite{leasure_national_2020} used a hierarchical Bayesian model to estimate population at \SI{100}{m} resolution grid cells nationally in Nigeria, and focused on ``provid[ing] robust probabilistic estimates of uncertainty at any spatial scale" \cite[p.~4]{leasure_national_2020}. 
The authors used the same settlement map as Weber et al.\cite{weber2018census} to remove unsettled grid cells prior to population density estimation, and  used additional geospatial covariates, including school density, average household size, and WorldPop gridded population estimates. WorldPop population estimates were generated using a census-dependent approach, so the proposed method in some sense integrates information from census into otherwise census-independent population predictions. The predicted population estimates, however, were not highly correlated with the true population counts \cite[Table 3]{leasure_national_2020}.

\subsection{Deep Learning for Intercensal Population Estimation}
\label{sec:deeppop}

There are several recent approaches that apply deep learning methods to intercensal population estimation using free and readily available high-resolution satellite imagery as opposed to relatively expensive very-high resolution imagery, and census as opposed to microcensus, potentially due to the prohibitive cost of collecting sufficient microcensus for training a custom deep neural network from scratch. HRSL uses very-high resolution imagery to focus on building footprint identification using a feedforward neural network and weakly supervised learning, and  redistributes the census proportionally to the fraction of built-up area \cite{tiecke_mapping_2017}, but does not use census as the response variable.

\textbf{Doupe et al.} \cite{doupe2016equitable} used an adapted VGG
\cite{simonyan15very} convolutional neural network (CNN) trained on a concatenation of low resolution Landsat-7 satellite images (7 channels) and night-time light images (1 channel). The VGG network was trained on observations generated from 18,421 population labeled enumeration areas from the 2002 Tanzanian census, and validated on observations generated from 7,150 labeled areas from the 2009 Kenyan census. The authors proposed using the output of the VGG network as a weighted surface for population disaggregation from regional population totals. This approach significantly outperformed AsiaPop (a precursor to WorldPop) in RMSE, \%RMSE, and MAE evaluation metrics \cite[Table 1]{doupe2016equitable}.

\textbf{Robinson et al.} \cite{robinson2017deep} trained an adapted VGG \cite{simonyan15very} CNN on Landsat-7 imagery from the year 2000 and US data from the year 2004, and validated it on Landsat and data from the year 2010. The authors split the US into 15 regions, and trained a model for each with around $\sim$800,000 training samples in total. Instead of predicting population count directly, the authors classified image patches into log scale population bands, and determined final population count by the network output weighted average of band centroids. Existing approaches for projecting data outperformed the final network when validated against the 2010 US census \cite[Table 1]{robinson2017deep}
, however the fitted model displayed an understanding of population, evidenced through visualizing the images that produced the highest probabilities for each population band \cite[Figure 5]{robinson2017deep}. 

\textbf{Hu et al.} \cite{hu_mapping_2019} generated population density maps at the village level in rural India using a custom VGG CNN based end-to-end learning. The authors used freely available high-resolution Landsat-8 imagery (\SI{30}{\m} resolution, RGB channels only) and Sentinel-1 radar imagery (\SI{10}{\m} resolution, converted to RGB) images of villages as predictor, and respective population from the 2011 Indian census as response. The training set included 350,000 villages and validation set included 150,000 villages across 32 states, and the resulting model outperformed two previous deep learning based approaches\cite{robinson2017deep,doupe2016equitable}. The authors observed that the approach performed better at a coarser district level resolution than a finer village level resolution \cite[Table 2]{hu_mapping_2019}.

Both census-dependent and census-independent approaches have their advantages and drawbacks. While census-dependent estimation is cheaper to perform using existing data, the results can be misleading if the  projected intercensal population count is inaccurate, and due to the limited resolution of both data and publicly available satellite imagery, these approaches exclusively predict population at a coarser spatial resolution.
 Census-independent estimation uses microcensus, which can be collected more frequently and is available at a finer scale, and although this data can be relatively expensive to collect in large enough quantities, it provides `ground truth' information at a finer scale which is not available for census-dependent approaches. 

\section{Methods and Data}
\label{sec:datamethods}

In this section we discuss some recent advancements in the principles and tools for self-supervised learning, partly in the context of remote sensing (\S \ref{sec:rep_transfer}), provide details of SCIPE, and the datasets used, i.e., satellite imagery and microcensus.

\subsection{Representation and Transfer Learning}
\label{sec:rep_transfer}

\begin{figure}[t]
        \centering
        \includegraphics[width=\linewidth]{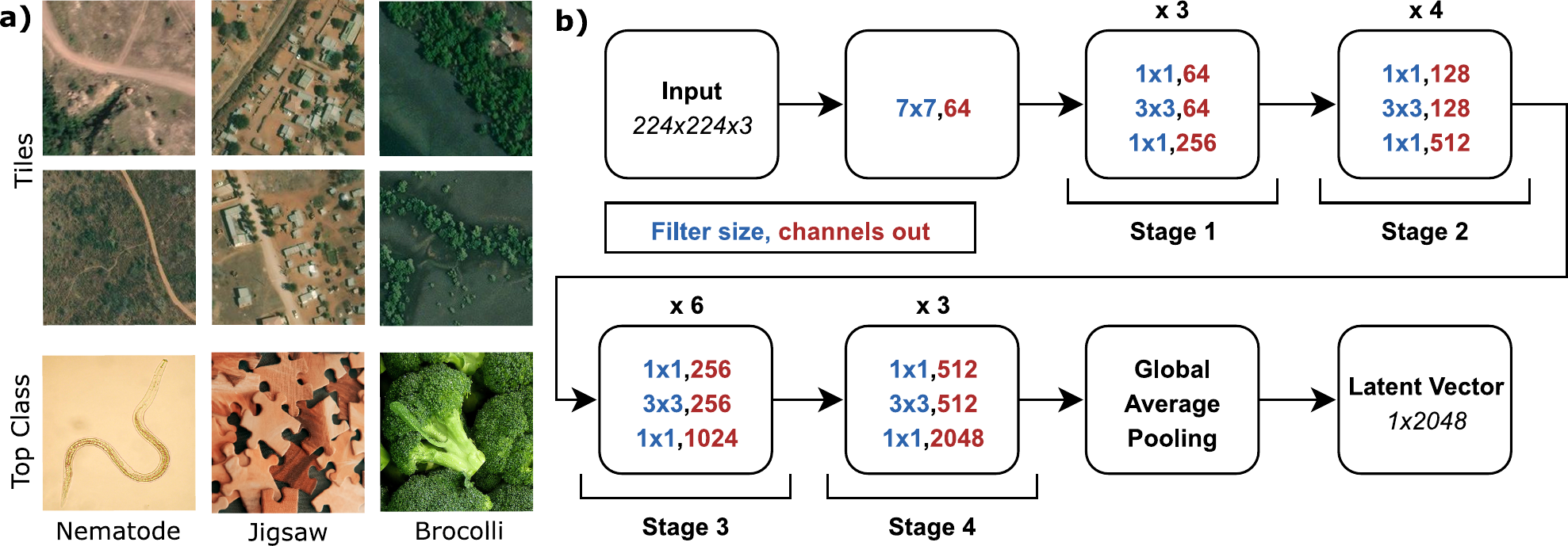}
        \caption{ResNet-50 ImageNet predictions and architecture. \textbf{a)} ImageNet class predictions for very-high-resolution satellite image tiles. Although the classes are irrelevant, the network shows an intrinsic understanding of the difference between built-up area, vegetation, and road. \textbf{b)} Diagram of ResNet-50 encoder. Residual blocks between Input and Global Average Pooling are comprised of convolutional layers with interleaved batch normalization layers.}
        \label{fig:resnet}
\end{figure}

Representation Learning learns a vector \emph{representation} of an image by transforming the image, for example, using deep neural network, such that the corresponding representation can be used for other tasks such as regression or classification using existing tools  \cite{bengio_representation_2013}. The learned representation can be used for transfer learning, i.e., using the transformation learned from a separate task, e.g., ImageNet classification, for a different one, e.g., population estimation \cite{razavian_cnn_2014}. Intuitively, this happens since a pre-trained network, although designed for a separate task, can extract meaningful features that can be informative for population estimation (see for example Figure~\ref{fig:resnet}a).

\textbf{Supervised pre-training} is a common approach for representation learning where a network is trained in a supervised learning context with a vast amount of annotated training data such as ImageNet. Once the network has been trained on this task, the output of the penultimate layer (or a different layer) of this pre-trained network can be used as a vector representation of the respective input image, and can be used as a predictor for further downstream analysis \cite{razavian_cnn_2014}. This approach works well in practice but its performance is inherently limited by the size of the dataset used for supervised learning which can be `small' \cite{caron2018deep}.
To mitigate this issue, representation learning using unsupervised methods such as Tile2Vec \cite{jean2019tile2vec}, and in particular, self-supervised approaches have become popular. Compared to supervised learning which maximizes the objective function of a pre-defined task such as classification, self-supervised learning generates pseudolabels from a pretext task in the absence of ground truth data, and different algorithms differ in the way they define the pretext tasks \cite{jing_self-supervised_2020}. 

In the context of population estimation, we focus on methods that either assume that the latent representations form clusters \cite{caron2018deep} or make them invariant to certain class of distortions \cite{zbontar2021barlow}. Our intuition is that grid tiles can be grouped together based on population. This is a common practice in census-independent population estimation, i.e., to split regions in categories and model these categories separately, e.g., see  \cite{leasure_national_2020} and \cite{doupe2016equitable}. We also observe this pattern in the representation space where built-up area separates well from uninhabited regions (see Figure~\ref{fig:ram}b). Additionally, we expect the population count of a grid tile to remain unchanged even if, for example, it is rotated, converted to grayscale, or resized.

\textbf{DeepCluster \cite{caron2018deep}} (and DeepClusterV2) jointly learns parameters of a deep neural network and the cluster assignments of its representations. It works by iteratively generating representations using an \emph{encoder} model that transforms the image to a vector, clustering these representations using a $k$-means clustering algorithm, and using these cluster assignments as pseudo-labels to retrain the encoder. The intuition behind this being that CNNs, even without training, can find partially useful representations, and therefore clusters, due to their convolutional structure. This weak signal can be exploited to bootstrap a process of iterative improvements to the representation and cluster quality.

\textbf{SwAV \cite{caron2020unsupervised}} clusters the representation while simultaneously enforcing consistency between cluster assignments produced for different distortions. This involves applying augmentations to each image, yielding two different \textit{views} of the image, which are then fed through the model, clustered, and compared to train the model. In particular, to enforce consistent cluster assignments between the views of the image, the cluster assignment of a view is predicted from the representation of another view of the same image. SwAV applies horizontal flips, color distortion and Gaussian blur after randomly cropping and resizing the image \cite[p.~15]{caron2020unsupervised}. 
Cropping affects the population of a grid tile, however, since satellite imagery alone can estimate population with some uncertainty, we assume that cropping will change population within this level of uncertainty. Although cropping is used as a data augmentation step in the existing pre-trained network, we avoid cropping as data augmentation when fine-tuning the network to predict population in \S\ref{sec:training}.

\textbf{Barlow Twins \cite{zbontar2021barlow}} also works by applying augmentations to each image, yielding two \textit{views} of the image, which are then fed through the model to produce two representations of the original image. To avoid  trivial constant solutions of existing self-supervised learning approaches aiming to achieve invariance to distortions, Barlow Twins considers a \emph{redundancy-reduction} approach, i.e., the network is optimized by maximizing the correlation along the main diagonal of the cross correlation matrix in the representation space to achieve invariance, while minimizing the correlation in all other positions to reduce redundancy of representations. Barlow Twins applies cropping, resizing, horizontal flipping, color jittering, converting to grayscale, Gaussian blurring, and solarization as distortions \cite[p.~3]{zbontar2021barlow}.

\subsection{Satellite Imagery and microcensus}
\label{sec:data}

\textbf{Satellite Imagery}
We used proprietary \SI{50}{cm} resolution satellite imagery (Vivid 2.0 from Maxar, WorldView-2 satellite) covering \SI{7773}{km^2} across two districts in Mozambique: Boane (BOA) and Magude (MGD). The Vivid 2.0 data product is intended as a base map for analysis. It has worldwide coverage and is updated annually with priority given to the low cloud coverage images, and hence images can be from different time periods and different sensors in the Maxar constellation. The product is provided already mosaicked and colour-balanced, increasing the transferability of any methods/algorithms developed using this data. The data are provided in a three-band combination of red, green and blue. The NIR band is not provided as part of the VIVID 2.0 data product. The procured data was a mosaic of images, mostly from 2018 and 2019 (83\% and 17\% for BOA and 43\% and 33\% for MGD, remainder from 2011 to 2020).

\textbf{Microcensus}
We used microcensus from 2019 conducted by SpaceSUR and GroundWork in these two districts, funded by UNICEF. The survey was conducted at a household level (with respective GPS locations available), and households were exhaustively sampled over several primary sampling units (PSUs) where PSUs were defined using natural boundaries such as road. We aggregated the household survey data to a \SI{100}{m} grid to generate population counts producing 474 labelled grid tiles. 

\textbf{Non-representative tiles}
Since the imagery and microcensus were not perfectly aligned temporally, and the PSUs had natural boundaries, many tiles contained either unsurveyed buildings or surveyed buildings absent in the imagery. Thus, the dataset contained both developed tiles (i.e. with many buildings) labeled as low population, or undeveloped tiles labeled as high population. Although such `outlier' tiles can be addressed with robust training, they cannot be used for validation. We, therefore, manually examined each grid tile by comparing the GPS location of surveyed buildings with those appearing in the imagery, and excluded those with a mismatch, leaving 199 curated tiles (CT).

\begin{figure}[tp]
    \centering
    \includegraphics[width=0.95\linewidth]{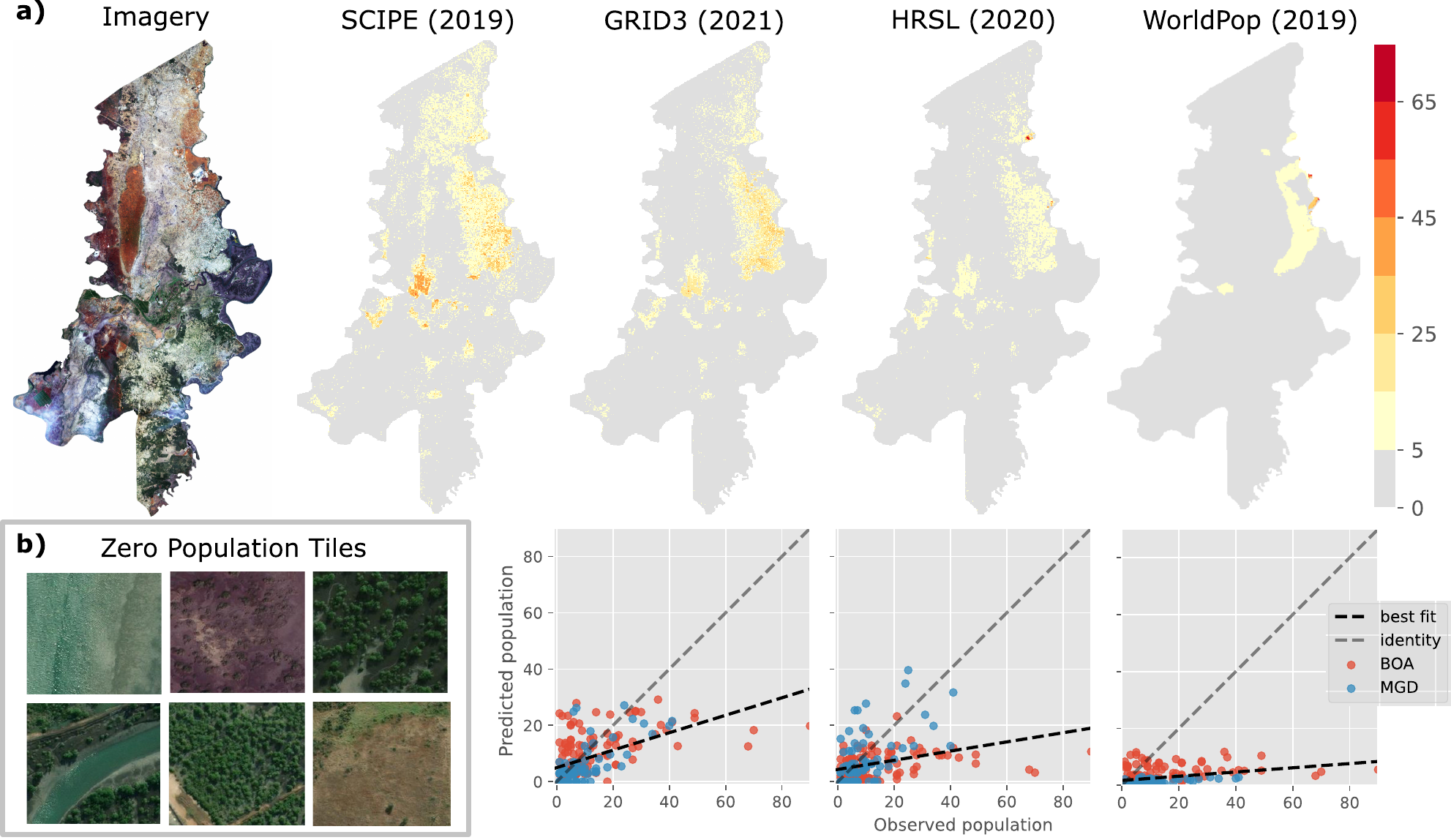}
    \caption{Regional population map comparison and zero population tiles. \textbf{a)} Gridded population count estimates over Boane (top), and comparison against microcensus (bottom). Year indicates target year for estimation where known, otherwise when estimates were published. We do not compare our results on the microcensus as this was our training and validation data. \textbf{b)} Examples of zero population tiles from \S\ref{sec:data}.}.
    \label{fig:main}
    \vspace{-1em}
\end{figure}

\textbf{Zero-population tiles}
Since the microcensus was conducted in settled areas, we had no labels for uninhabited tiles. Although this does not pose a problem when comparing the performance of different models on the available microcensus (Table~\ref{tab:deeplearningpop}), the models do not learn to predict zero population when applied to an entire district which will include many uninhabitated areas. To resolve this, we identified 75 random tiles (50 from BOA, 25 from MGD) with zero population (ZT) guided by HRSL, i.e., from regions where HRSL showed zero population. We selected more ZTs from BOA to improve regional population estimates (see Figure~\ref{fig:main}b). Thus, we had 274 grid tiles in total.

\subsection{Models and Training} \label{sec:training}

We use a ResNet-50 \cite{he2016deep} CNN architecture to estimate population from grid tiles. The model architecture is shown in Figure~\ref{fig:resnet}b with  $224\times224\times3$ dimensional input and 49 convolutional layers
followed by a Global Average Pooling layer which results in a 2048 dimensional latent representation.
We used the pre-trained ResNet-50 models trained on ImageNet using methods described in \S\textbf{\ref{sec:rep_transfer}} after
resizing the grid tiles of size $200\times 200\times 3$ (\SI{100}{\m} RGB) to $224\times 224\times 3$, and used these (representation, population) pairs to train a prediction model using Random Forest. The hyperparameters of the model were chosen using a grid search over \texttt{num\_estimators} $\in\{100,200,...,500\}$, \texttt{min\_samples\_split} $\in\{2,5\}$ and \texttt{min\_samples\_leaf} $\in\{1,2\}$. A linear regression head can also be trained to predict population in an end-to-end manner. This yields several advantages: rapid inference on GPUs, a simple pipeline, and a simple method for determining uncertainty. However, we observed that the Random Forest model outperformed the linear regression head.

\textbf{Pre-trained model} We used pre-trained ResNet50 models described in \S\ref{sec:rep_transfer} to extract representations, and also fine-tuned these models with microcensus.

\textbf{Fine-tuning}
We fine-tuned the pretrained models using a combination of curated and zero grid tiles after attaching a \emph{linear regression head} following the global average pooling layer and minimizing the $\ell_2$ loss between observed and predicted population.
Given the labelled grid tiles (number of grid tiles vary depending on experimental set-up), we randomly split them into training and validation sets (80-20\%). Due to the limited number of tiles in the dataset, we apply random dihedral transformations (i.e. reflections and rotations) to tiles to augment the training set, avoiding transformations that could affect the validity of the population count e.g. crops that could remove buildings. We use Adam optimizer to minimize the loss function which takes about $1$ minute with a batch size of $32$ on a single Nvidia GTX 1070 with \SI{8}{GB} of VRAM. During training, first, the network was frozen (i.e., the weights were kept fixed) and only the regression head was trained for 5 epochs with a learning rate of $2\times10^{-3}$, and second, the entire network was trained using a \textit{discriminative learning rate} \cite{howard2020deep}, where the learning rate is large at the top of the network, and is reduced in the earlier layers, avoiding large changes to the earlier layers of the model which typically extract more general features, and focusing training on the domain-specific later layers. The base learning rate at the top of the network was $1\times10^{-3}$, and it was decreased in the preceding stages to a minimum of $1\times10^{-5}$. We used early stopping to halt training when validation loss plateaued (i.e., no improvement for 2 or more epochs) to avoid overfitting.

\begin{figure}[t]
    \begin{minipage}[t]{.5\textwidth}
        \vspace{0pt}
        \centering
        \resizebox{\linewidth}{!}{
        \begin{tabular}{lccccc}
            \toprule
            Features used & \multicolumn{1}{c}{$R^2$} & \multicolumn{1}{c}{$\MeAPE$} & \multicolumn{1}{c}{$\MAE$} & \multicolumn{1}{c}{$\textsc{IQR}$} & \multicolumn{1}{c}{$\AggPE$} \\ 
            \cmidrule(l){1-1} \cmidrule(l){2-6}
            \addlinespace
            \textit{Hand-crafted features:} \\
            \rowcolor{Gray}
            Public Only & -0.22 &  57.8\% & 5.05 & 8.60 &  21.5\% \\
            Footprint Only &  \textbf{0.47} &  \textbf{44.5\%} & \textbf{3.75} & 5.86 &  \textbf{02.2\%} \\
            \rowcolor{Gray}
            Public + Footprint & 0.46 &  48.8\% & 4.63 & 5.11 &  07.6\% \\
            \addlinespace
            \textit{Representation Learning:} \\
            \rowcolor{Gray}
            Supervised &  0.20 &  54.7\% & 5.36  & 7.97 &  02.0\% \\
            Supervised (FT) & 0.33 &  52.9\% &  4.72 & 6.23  &  05.5\% \\
            \rowcolor{Gray}
            SWAV  & 0.34 &  51.6\% & 6.60 &  4.33 &  00.8\% \\
            SWAV (FT) &  \textbf{0.41} &  46.9\% &  5.83 &  4.35 &  03.5\% \\
            \rowcolor{Gray}
            DeepCluster  & 0.26 &  50.3\% & 4.60 &  6.03 & 06.7\% \\
            DeepCluster (FT) & 0.13 &  62.5\% & 5.98 & 8.32 & 06.8\% \\
            \rowcolor{Gray}
            Barlow Twins & 0.27 &  51.9\% & 5.40  & 6.65  &  02.8\% \\
            Barlow Twins (FT) & 0.39 &  \textbf{44.0\%} &  \textbf{3.91} & 6.32 & \textbf{01.1\%} \\
            \addlinespace
            \textit{Null Model:} \\
            \rowcolor{Gray}
            None & -0.12 & 76.45\% & 7.57 & 10.0 & 01.7\% \\
            \cmidrule(l){1-6}\addlinespace
            Existing Maps \\
            \cmidrule(l){1-6}
            \rowcolor{Gray}
            GRID3 & \textbf{0.22} & \textbf{51.7\%} & \textbf{4.25} & 7.11 & \textbf{26.7\%} \\
            HRSL & -0.12 & 70.7\% & 5.04 & 7.94 & 46.8\% \\
            \rowcolor{Gray}
            WorldPop & -0.41 & 86.8\% & 5.85 & 8.18 & 77.9\% \\
            \bottomrule
        \end{tabular}
        }
    \captionsetup{width=0.95\linewidth}
    \captionof{table}{Population model validation performance for (\textbf{top}) manually and automatically extracted features, along with (\textbf{bottom}) test performance of existing maps. See \S \textbf{\ref{sec:evaluation}} for metric definitions. \textsc{IQR} is interquartile range of absolute errors. \textbf{Bold} indicates best in category. FT is fine-tuning.}
    \label{tab:results}
  \end{minipage}
  \begin{minipage}[t]{0.48\textwidth}
  \vspace{0pt}
    \centering
    \includegraphics[width=\linewidth]{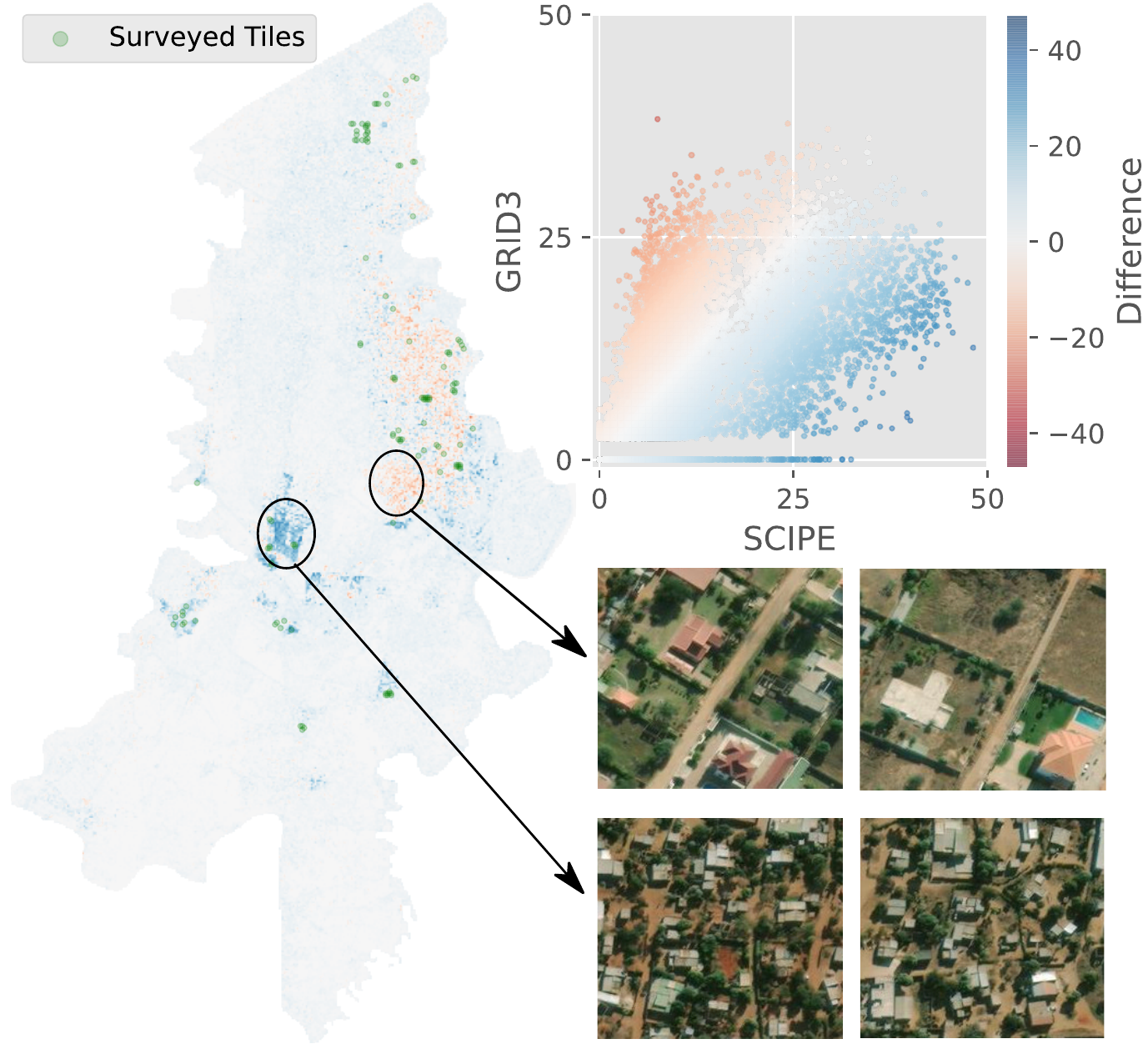}
    \captionsetup{width=0.95\linewidth}
    \captionof{figure}{Clockwise from left: Difference between SCIPE and GRID3 population maps for Boane, the respective scatter plot, and examples of survey tiles from regions where they differ.}
    \label{fig:grid3}
  \end{minipage}
\end{figure}

\subsection{Evaluation Metrics and Cross-validation}
\label{sec:evaluation}

\textbf{Evaluation Metrics}
We compare the different methods against several evaluation metrics, i.e., R-squared, median absolute error (\MAE), median absolute percentage error (\MeAPE), and aggregated percentage error (\AggPE) (to capture average error at regional levels characterised by $A$), as follows,
\begin{align*}
&R^2 = 1- {\sum_i (y_i - \hat{y})^2}/{\sum_i (y_i - \bar{y})^2} & 
\MAE &= \textrm{median}\, |y_i - \hat{y}_i| \\
&\MeAPE = \textrm{median}\, {|y_i - \hat{y}_i|}/{y_i} &
\AggPE &= \textrm{median}_A \, {\left|\sum_{i \in A} y_i - \sum_{i \in A} \hat{y}_i\right|}/{\sum_{i \in A} y_i}
\end{align*}
These evaluation metrics capture different aspects of the prediction, and each has different significance. For example, $R^2$ may be dominated by large population counts while {\MeAPE} may be dominated by small population counts.

\textbf{Null Model}
As `null model' we predict population as the mean of the training set irrespective of feature values. We used this as an initial baseline to ensure any perceived performance when transferring features from ImageNet is not trivial. 

\textbf{Baseline}
To properly assess the performance of automatic feature extraction, we compared to results when using hand-crafted features and public datasets to predict population that is more common for census-independent population estimation. We took a variety of public features (Landsat imagery, land cover classification, OSM road data, night-time lights), along with building footprints automatically extracted from each image tile using a U-Net model pre-trained on SpaceNet and fine-tuned with `dot-annotation' from non-surveyed buildings, and using these features to train a Random Forest model. 

\textbf{Cross-validation}
We compare the different approaches to population estimation using cross-validation.
For each region, we partitioned the data into four subsets spatially, and formed validation folds by taking the union of these subsets across the two regions. We reported the evaluation metrics over \emph{pooled} predictions from the four validation folds covering the entire microcensus. When fine-tuning, we trained one network for each fold separately (to avoid data leakage) resulting in four networks.

\section{Results}
\label{sec:evaluation}
In this section, we compare the performance of several self-supervised learning frameworks using cross-validation, apply the best performing model to predict a regional map of Boane, and compare it against existing maps from GRID3, HRSL and WorldPop where all of these methods take a census-dependent approach to population estimation within our ROIs, and cannot be regarded as `ground-truth'.  We show the interpretability of the framework using uncertainty quantification and activation maps.

\textbf{Model selection}
Table \ref{tab:results} shows the cross-validation results for population estimation using Random Forest regression on representations extracted using ResNet-50 model with curated tiles only. We observe that, 1) representations extracted using any pre-trained network outperformed estimation using publicly available features in all but $\MAE$ metric,  2) fine-tuning any of the representation learning frameworks with microcensus, besides DeepCluster, resulted in an improvement of the performance of the framework, 3) although all of the representation learning framework (best $\MAE=3.91$) outperformed than the null model ($\MAE=7.57$), the baseline models trained with building footprint area (best $\MAE=3.75$) as a feature still outperformed them in $R^2$ and $\MAE$, and 4). Barlow Twins overall had lower error metrics and the second largest $R^2$ metric among the fine-tuned models, so we consider this model for further analysis. We did not evaluate the performance of Tile2Vec since the available pre-trained model required an NIR band for input data, which Vivid 2.0 lacks.

\textbf{Regional Population Estimation}
We use representations learned with a ResNet50 architecture pre-trained on ImageNet using Barlow Twins and fine-tuned using curated and zero grid tiles (with 80-20 random train/validation split of all tiles, no cross-validation) to extract representations for our survey tiles. These representations are used to train a Random Forest model (\S\ref{sec:training}), which is used to produce a population map for Boane district. The map is shown in Figure \ref{fig:main}a along with three existing population maps from WorldPop, HRSL and GRID3.
We observe that, with respect to our `ground truth' microcensus, 1) GRID3 provides a more accurate population map of Boane than HRSL and WorldPop, but usually underestimates population, 2) WorldPop lacks the finer details of the other population maps, and underestimates population, 3) although the settlement map provided by HRSL matches that of SCIPE and GRID3 well, its similarity with SCIPE is less than that of GRID3, and 4) SCIPE and GRID3 provide visually similar settlement map, and the population estimates are also more similar in scale compared to HRSL and WorldPop. 

\textbf{Census} We additionally compare the aggregated population estimate in Boane with 2019 census projection from 2017 census, and we observe that SCIPE overestimates population by 29\%. Although projected census is not the ground truth, the discrepancy in estimated population is potentially due to SCIPE not modelling zero population explicitly. SCIPE can be extended by using, for example, a zero-inflated population model to model the zero-population better. We leave this as future work. 

Since GRID3 provides a more accurate population map than HRSL and WorldPop, we compare it against SCIPE in more detail. Figure~\ref{fig:grid3} shows the difference in population maps produced by these two approaches. We observe that, 1) the estimated population of these approaches matched well quantitatively (Spearman's $\rho$ 0.70, Pearson's $\rho$ 0.79), 
2) there are regions where SCIPE underestimated population, and these are areas where microcensus was not available, and 3) there are regions where GRID3 underestimated population, and they usually coincided with regions where microcensus was available and SCIPE could potentially provide better estimates. Therefore, there is a high level of agreement between the two products and they provide similar estimates, and discrepancies appear in regions that lack microcensus for training. A more detailed comparison of these two population maps will be valuable, and may lead to both improved population estimation through ensemble learning and better microcensus data collection through resolving model disagreements. However, this is beyond the scope of this work.

\begin{figure}[tbp!]
    \centering
    \includegraphics[width=\linewidth]{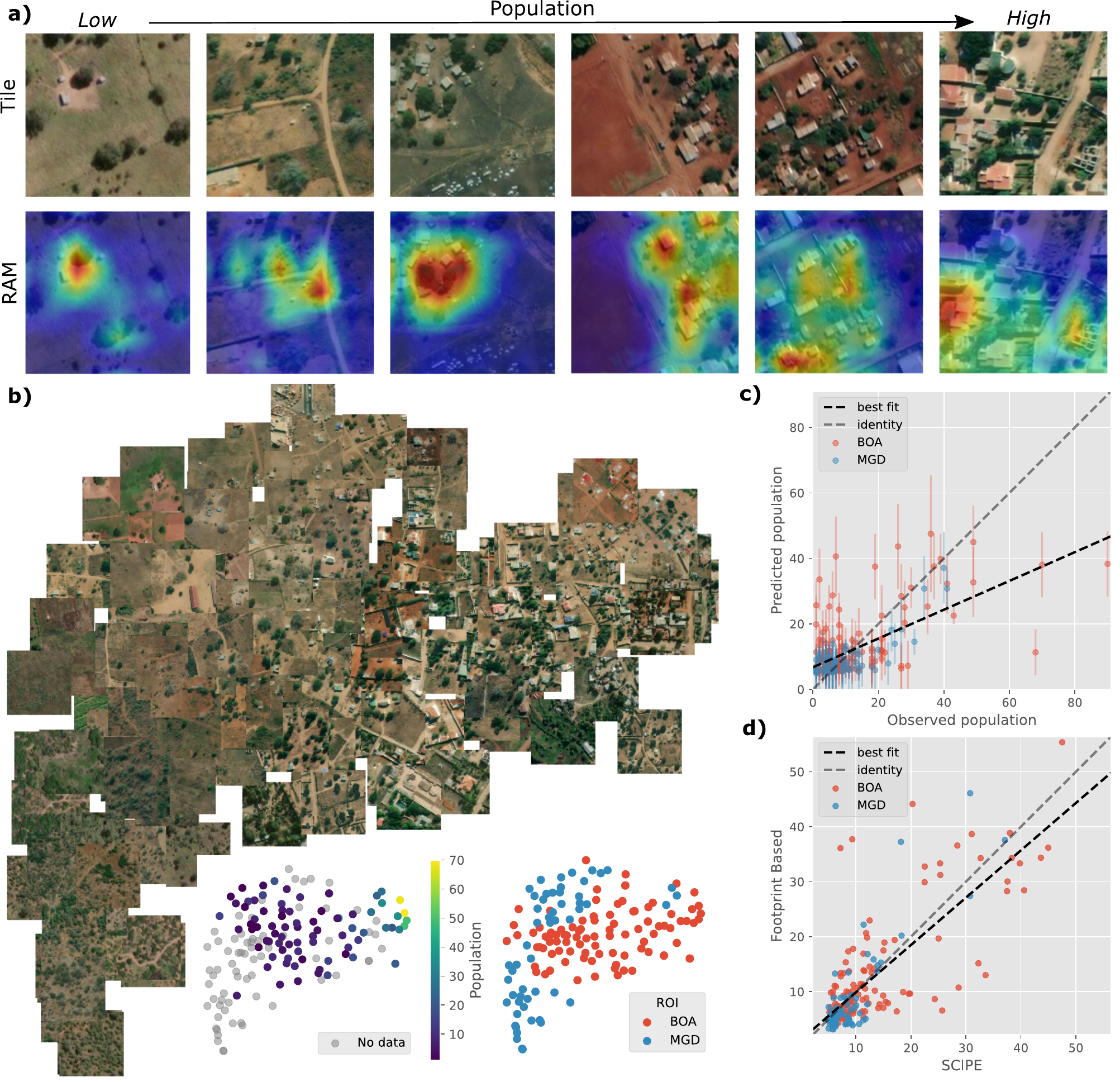}
    \caption{Activation maps, embedding visualization, and comparison of SCIPE with microcensus data and footprint based estimates. \textbf{a)} Tiles and their associated regression activation maps (RAMs) from fine-tuned Barlow Twins model \textbf{b)} Plot of t-SNE embeddings of representations from Barlow Twins for 75 random tiles and 75 microcensus tiles from Boane and Magude (main), with embeddings coloured by population (bottom left) and region (bottom right) \textbf{c)} Predicted over observed plot over microcensus with prediction uncertainty \textbf{d)} Comparison between SCIPE estimates and building footprint based estimates on the microcensus grid tiles.}
    \label{fig:ram}
\end{figure}

\textbf{Uncertainty}
To further assess the quality of the estimated population map, we quantify the uncertainty and qualitatively assess their `explanations'. We can assess the uncertainty of predictions in several ways either at the level of the representation learning or at the level of the Random Forest population model. For the former, we can apply Monte Carlo dropout \cite{gal2016dropout} by placing dropout layers ($p=0.1$) after each stage of the ResNet models, and predicting multiple population value for each grid tile.
For the latter, the uncertainty can be quantified from the output of the individual decision trees in the Random Forest model without perturbing the representation. 
Figure \ref{fig:ram}c shows Random Forest model predictions on fine-tuned Barlow Twins features and their associated uncertainties. We observe that the estimated uncertainty matches the intuition of higher estimated population having higher uncertainty.

\textbf{Explanation}
To assess the outcome of the model, we use regression activation maps (RAMs) \cite{wang2017diabetic} that show the discriminative regions of input image that is informative of the outcome of the model. It is widely reported that building footprint area is an important indicator of population, and we observe that a fine-tuned Barlow Twins model produces RAM plots that show a clear focus on built-up area, which agrees with the intuition (see Figure~\ref{fig:ram}a). To further explore if SCIPE focuses on built-up area to estimate population, we observe that population estimates using SCIPE and that using building footprints (as presented in Table~\ref{tab:results}) show high correlation (Spearman's $\rho$ 0.68, Pearson's $\rho$ 0.74 over Boane region) (see Figure~\ref{fig:ram}d) corroborating this observation. 

\textbf{Embedding} Finally, we visualize the representations available from the fine-tuned Barlow Twins model to assess if they meaningfully separate in terms of population, and we observe that this is indeed the situation (see Figure~\ref{fig:ram}b). 

\section{Discussion}
\label{sec:discussion}

We find representation learning to be an effective tool for estimating population at a medium-resolution from limited local microcensus. Although this approach did not outperform building footprint area based estimations; it is fast, does not require human supervision, and only relies on very-high-resolution satellite images, making it sustainable and transferable in the sense that users can extrapolate their own local microcensus with relative ease, and also quantify uncertainty and capture explanations.

There is likely a hard limit to the predictive power of satellite imagery alone, owed to the difficulty of distinguishing inhabited and uninhabited areas in some contexts. For example, Robinson et al.\cite{robinson2017deep} gave the example of Walt Disney World which is built to look like a settled area but has $0$ population. To address this issue, an interesting extension of SCIPE will be to use multiple data sources, such as night-time light, land-cover data, altitude and slope information, location of services, of possibly different resolutions in the model alongside very-high-resolution imagery without changing its core focus, i.e., of using pre-trained network and fine-tuning them with limited amount of microcensus. This can potentially improve the prediction of population in areas that are uninhabited. We have also used a Random Forest model which effectively treats grid tiles as independent and identically distributed samples which they are not. Considering the spatial arrangement of grid tiles can improve estimation further by using broader contextual information around it, for example to establish the socioeconomic status or land use of the surrounding areas.

Ideally, we would like to use self-supervised learning framework directly on satellite images to learn appropriate representation, rather than relying on pre-trained networks and fine-tuning. This will, however, require a vast quantity of training data, and has not been the focus of this work. In this work we have focused on \emph{sustainability}, both in terms of human annotation and computational resources which prohibits training from scratch, and have shown that existing tools can be used to produce reliable population estimates. The proposed framework should also be validated externally on a larger scale. We explored population of a single district in Mozambique while existing population maps are available over the whole of Africa. Assessing the utility of SCIPE better would require further large scale validation both in different regions of Mozambique (which is our immediate focus), and in other countries (which is our long term goal), to make population maps more frequent, accessible, reliable and reproducible.

SCIPE avoids several typical bottlenecks associated with census-independent population estimation. While some methods require tedious manual annotation of built up area or potentially incomplete public features, SCIPE extracts features automatically using only satellite images. SCIPE is extremely fast, requires negligible GPU time, and provides meaningful population estimates. Microcensus data may not be available in all countries or regions, and can be expensive to gather, but this cost is far lower than that of conducting census on a regular basis. Very-high-resolution satellite imagery can also be expensive, but has become more accessible in recent years when used for humanitarian purposes \cite[p.~14]{engstrom_estimating_2020}. Given that many development agencies benefit from subsidised access to the Maxar very-high-resolution imagery these population maps could be produced relatively quickly for specific regions of focus, for example, when vaccination programmes are being planned. This approach, therefore, would contribute towards the UNs stated need for a data revolution \cite{world2014counts} by allowing regularly updated estimates of population between census enumeration periods supporting a range of humanitarian activities as well as general governmental and NGO planning and allocation of resources.

\bibliographystyle{plain}
\bibliography{population}

\begin{thebibliography}{10}

\bibitem{world2014counts}
Independent expert advisory group on a data revolution for sustainable
  development. {A} world that counts: Mobilising the data revolution for
  sustainable development.
\newblock November 2014.

\bibitem{bengio_representation_2013}
Yoshua Bengio, Aaron Courville, and Pascal Vincent.
\newblock Representation learning: A review and new perspectives.
\newblock 2013.

\bibitem{soton445265}
Maksym Bondarenko, Patricia Jones, Douglas Leasure, Attila Lazar, and Andrew
  Tatem.
\newblock Gridded population estimates disaggregated from mozambique's fourth
  general population and housing census (2017 census), version 1.1, November
  2020.

\bibitem{caron2018deep}
Mathilde Caron, Piotr Bojanowski, Armand Joulin, and Matthijs Douze.
\newblock Deep clustering for unsupervised learning of visual features.
\newblock In {\em Proceedings of the ECCV}, 2018.

\bibitem{caron2020unsupervised}
Mathilde Caron, Ishan Misra, Julien Mairal, Priya Goyal, Piotr Bojanowski, and
  Armand Joulin.
\newblock Unsupervised learning of visual features by contrasting cluster
  assignments.
\newblock In {\em Proceedings of NeurIPS}, 2020.

\bibitem{doupe2016equitable}
Patrick Doupe, Emilie Bruzelius, James Faghmous, and Samuel~G Ruchman.
\newblock Equitable development through deep learning: The case of sub-national
  population density estimation.
\newblock In {\em Proceedings of {ACM} {DEV}}, pages 1--10, 2016.

\bibitem{engstrom_estimating_2020}
Ryan Engstrom, David Newhouse, and Vidhya Soundararajan.
\newblock Estimating small-area population density in {Sri} {Lanka} using
  surveys and {Geo}-spatial data.
\newblock {\em {PLoS} {ONE}}, 15(8):e0237063, August 2020.

\bibitem{gal2016dropout}
Yarin Gal and Zoubin Ghahramani.
\newblock Dropout as a bayesian approximation: Representing model uncertainty
  in deep learning.
\newblock In {\em Proceedings of ICML}, pages 1050--1059. PMLR, 2016.

\bibitem{he2016deep}
Kaiming He, Xiangyu Zhang, Shaoqing Ren, and Jian Sun.
\newblock Deep residual learning for image recognition.
\newblock In {\em Proceedings of CVPR}, 2016.

\bibitem{hillson_estimating_2019}
Roger Hillson, Austin Coates, Joel~D. Alejandre, Kathryn~H. Jacobsen, Rashid
  Ansumana, Alfred~S. Bockarie, Umaru Bangura, Joseph~M. Lamin, and David~A.
  Stenger.
\newblock Estimating the size of urban populations using {Landsat} images: a
  case study of {Bo}, {Sierra} {Leone}, {West} {Africa}.
\newblock {\em Int. J. Health Geogr.}, 2019.

\bibitem{howard2020deep}
Jeremy Howard and Sylvain Gugger.
\newblock {\em Deep Learning for Coders with fastai and PyTorch}.
\newblock O'Reilly Media, 2020.

\bibitem{hu_mapping_2019}
Wenjie Hu, Jay~Harshadbhai Patel, Zoe-Alanah Robert, Paul Novosad, Samuel
  Asher, Zhongyi Tang, Marshall Burke, David Lobell, and Stefano Ermon.
\newblock Mapping missing population in rural india: A deep learning approach
  with satellite imagery.
\newblock 2019.

\bibitem{jean2019tile2vec}
Neal Jean, Sherrie Wang, Anshul Samar, George Azzari, David Lobell, and Stefano
  Ermon.
\newblock Tile2vec: Unsupervised representation learning for spatially
  distributed data.
\newblock In {\em Proceedings of the AAAI}, 2019.

\bibitem{jing_self-supervised_2020}
Longlong Jing and Yingli Tian.
\newblock Self-supervised {Visual} {Feature} {Learning} with {Deep} {Neural}
  {Networks}: {A} {Survey}.
\newblock {\em IEEE Trans. Pattern Anal. Mach. Intell.}, 2020.

\bibitem{hrsl}
Facebook~Connectivity Lab and Center for International Earth Science
  Information Network CIESIN Columbia~University.
\newblock High resolution settlement layer {(HRSL)}, 2016.

\bibitem{leasure_national_2020}
Douglas~R. Leasure, Warren~C. Jochem, Eric~M. Weber, Vincent Seaman, and
  Andrew~J. Tatem.
\newblock National population mapping from sparse survey data: a hierarchical
  bayesian modeling framework to account for uncertainty.
\newblock {\em {PNAS}}, 2020.

\bibitem{linard2012population}
Catherine Linard, Marius Gilbert, Robert~W Snow, Abdisalan~M Noor, and Andrew~J
  Tatem.
\newblock Population distribution, settlement patterns and accessibility across
  africa in 2010.
\newblock {\em {PLoS} {ONE}}, 2012.

\bibitem{linardatos_explainable_2020}
Pantelis Linardatos, Vasilis Papastefanopoulos, and Sotiris Kotsiantis.
\newblock Explainable {AI}: {A} {Review} of {Machine} {Learning}
  {Interpretability} {Methods}.
\newblock {\em Entropy}, 2020.

\bibitem{razavian_cnn_2014}
Ali~Sharif Razavian, Hossein Azizpour, Josephine Sullivan, and Stefan Carlsson.
\newblock {CNN} features off-the-shelf: An astounding baseline for recognition.
\newblock In {\em Proceedings of CVPR Workshops}, pages 512--519, 2014.

\bibitem{robinson2017deep}
Caleb Robinson, Fred Hohman, and Bistra Dilkina.
\newblock A deep learning approach for population estimation from satellite
  imagery.
\newblock In {\em Proceedings of ACM SIGSPATIAL Workshop on Geospatial
  Humanities}, 2017.

\bibitem{Shearmur10editorial}
Richard Shearmur.
\newblock Editorial - {A} world without data? {The} unintended consequences of
  fashion in geography.
\newblock {\em Urban Geography}, 2010.

\bibitem{simonyan15very}
Karen Simonyan and Andrew Zisserman.
\newblock Very deep convolutional networks for large-scale image recognition.
\newblock In {\em Proceedings of ICLR}, 2015.

\bibitem{tiecke_mapping_2017}
Tobias~G. Tiecke, Xianming Liu, Amy Zhang, Andreas Gros, Nan Li, Gregory
  Yetman, Talip Kilic, Siobhan Murray, Brian Blankespoor, Espen~B. Prydz, and
  Hai-Anh~H. Dang.
\newblock Mapping the world population one building at a time.
\newblock 2017.
\newblock arXiv: 1712.05839.

\bibitem{wang2017diabetic}
Zhiguang Wang and Jianbo Yang.
\newblock Diabetic retinopathy detection via deep convolutional networks for
  discriminative localization and visual explanation.
\newblock In {\em Proceedings of AAAI Workshops}, 2018.

\bibitem{wardrop2018spatially}
NA~Wardrop, WC~Jochem, TJ~Bird, HR~Chamberlain, D~Clarke, D~Kerr, L~Bengtsson,
  S~Juran, V~Seaman, and AJ~Tatem.
\newblock Spatially disaggregated population estimates in the absence of
  national population and housing census data.
\newblock {\em PNAS}, 2018.

\bibitem{weber2018census}
Eric~M Weber, Vincent~Y Seaman, Robert~N Stewart, Tomas~J Bird, Andrew~J Tatem,
  Jacob~J McKee, Budhendra~L Bhaduri, Jessica~J Moehl, and Andrew~E Reith.
\newblock Census-independent population mapping in northern nigeria.
\newblock {\em Remote Sens. Environ}, 2018.

\bibitem{zbontar2021barlow}
Jure Zbontar, Li~Jing, Ishan Misra, Yann LeCun, and Stéphane Deny.
\newblock Barlow twins: Self-supervised learning via redundancy reduction,
  2021.

\end{thebibliography}

\subsection*{Acknowledgments}
The project was funded by the Data for Children Collaborative with UNICEF.

\end{document}